\documentclass[conference]{IEEEtran}
\usepackage{cite}
\usepackage{amsmath,amssymb,amsfonts}
\usepackage{algorithm}
\usepackage{algorithmic}
\usepackage{graphicx}
\usepackage{textcomp}
\usepackage{xcolor}
\usepackage{url}

\usepackage{multirow}
\usepackage{makecell}
\usepackage{threeparttable}
\usepackage{multirow,graphicx}
\usepackage[hidelinks]{hyperref}   
\renewcommand{\figurename}{Fig.}

\def\BibTeX{{\rm B\kern-.05em{\sc i\kern-.025em b}\kern-.08em
    T\kern-.1667em\lower.7ex\hbox{E}\kern-.125emX}}
\begin{document}

\title{Effective Series Decomposition and Components Learning for Time Series Generation}

\author{%
Zixuan Ma$^{1}$, Chenfeng Huang$^{2}$\\
$^{1}$School of Engineering and Applied Science, University of Pennsylvania, Philadelphia, USA\\
$^{2}$Viterbi School of Engineering, University of Southern California, Los Angeles, USA\\
\textit{\{zixuanm@seas.upenn.edu, chuang78@usc.edu\}}%
}

\maketitle

\begin{abstract}
Time series generation focuses on modeling the underlying data distribution and resampling to produce authentic time series data. Key components, such as trend and seasonality, drive temporal fluctuations, yet many existing approaches fail to employ interpretative decomposition methods, limiting their ability to synthesize meaningful trend and seasonal patterns. To address this gap, we introduce Seasonal-Trend Diffusion (STDiffusion), a novel framework for multivariate time series generation that integrates diffusion probabilistic models with advanced learnable series decomposition techniques, enhancing the interpretability of the generation process. Our approach separates the trend and seasonal learning into distinct blocks: a Multi-Layer Perceptron (MLP) structure captures the trend, while adaptive wavelet distillation facilitates effective multi-resolution learning of seasonal components. This decomposition improves the interpretability of the model on multiple scales. In addition, we designed a comprehensive correction mechanism aimed at ensuring that the generated components exhibit a high degree of internal consistency and preserve meaningful interrelationships with one another.

Our empirical studies on eight real-world datasets demonstrate that STDiffusion achieves state-of-the-art performance in time series generation tasks. Furthermore, we extend the model’s application to multi-window long-sequence time series generation, which delivered reliable results and highlighted its robustness and versatility. The source code of our model is officially released as STDiffusion on Github \url{https://github.com/mobkageyama/STDiffusion}.
\end{abstract}

\begin{IEEEkeywords}
time series generation; diffusion model; wavelet transformation; moving average
\end{IEEEkeywords}

\section{Introduction}
Time series data, which record sequential observations over time, are critical for decision-making in domains like finance, healthcare, and energy \cite{b1}\cite{b2}. Yet real-world datasets are often limited by issues such as data scarcity, missing values, or privacy concerns \cite{b3}. Generating realistic synthetic time series can help alleviate these constraints, enabling more robust model development and evaluation. However, the inherent sequential nature and temporal dependencies of time series data present unique challenges in creating such synthetic datasets \cite{b4}. Recognizing these challenges, Generative Adversarial Networks (GANs) and Variational Autoencoders (VAEs) have been widely applied to time series generation. However, GANs often suffer from training instability and mode collapse \cite{b7}\cite{b8}, while VAEs tend to produce overly smooth or blurry outputs due to posterior mismatch, limiting their effectiveness in capturing fine-grained temporal structures \cite{b19}.

To overcome these challenges, diffusion models have recently emerged as a powerful alternative. Originally successful in the domains of image, speech, and text generation \cite{b12}\cite{b13}\cite{b14}, diffusion models have demonstrated stronger compatibility and stability than traditional generative methods when modeling complex temporal dynamics \cite{b15}. Unlike GANs, diffusion models avoid mode collapse by learning to approximate the full data distribution via a gradual denoising process \cite{b18}. Compared to VAEs, they achieve sharper and more structured outputs without sacrificing probabilistic modeling.

In the context of time series, diffusion models have shown great potential in handling diverse tasks such as forecasting, imputation, and anomaly detection \cite{b16}\cite{b17}. Meanwhile, several advanced time series generation methods have been proposed, such as FTS-Diffusion\cite{b20}, which synthesizes financial time series by capturing irregular and scale-invariant patterns, and Diffusion-TS\cite{b21} integrate sequence decomposition in latent space via one-dimensional convolution to capture trend and seasonal components, improving the quality of the generation.

Although Diffusion-TS claims interpretability by applying decomposition frameworks in high-dimensional latent spaces, their reliance on neural network embeddings inherently limits transparency. It remains unclear whether the extracted components correspond to comprehensible structures such as trends and seasonality. In contrast, traditional time series decomposition methods operate directly on raw data, segmenting the series into additive or multiplicative components under fixed assumptions. These approaches effectively isolate trends, seasonality, and residuals without relying on learned high-dimensional representations \cite{b22}.

To bridge the gap between interpretability and the representational power of deep models, we propose STDiffusion, a novel diffusion-based framework that integrates the strengths of traditional decomposition with the flexibility of neural architectures. Specifically, we introduce a learnable trend-seasonal decomposition applied directly to raw inputs rather than latent embeddings. By employing a Learnable Moving Average (LMA) mechanism, STDiffusion extracts components with explicit semantic meaning that are retained and explicitly modeled throughout the generation process. Unlike black-box models such as Diffusion-TS, which perform repeated decomposition on hidden states, our method produces disentangled and interpretable signals in a single stage, improving both clarity and efficiency.

Although our decomposition strategy encourages statistical separation of trend and seasonal components, this typically holds only in a low-order sense, such as decorrelation in frequency or moving average domains. Real-world time series often exhibit complex high-order dependencies, including delayed effects or causal interactions across varying temporal resolutions \cite{b55}. To address this, we introduce a seasonal-trend correction module, enabling mutual conditioning between components. The correction mechanism provides consistent scaling across components, which is critical in high-dimensional settings to prevent signal collapse and instability during generation. Moreover,  this mechanism restores coherent joint dynamics that may otherwise be lost under marginally independent modeling assumptions, ensuring temporal consistency and scale alignment critical for generating stable and interpretable sequences.

The main contributions of our work can be summarized as follows:
\begin{itemize}
\item We propose \textbf{STDiffusion}, a diffusion-based generative framework that introduces component-wise interpretable modeling in time series generation, combining the strengths of traditional decomposition and modern deep learning.
\item We develop a \textbf{learnable moving average (LMA)} module that operates directly on raw input, enabling the data-adaptive extraction of trend and seasonal components with explicit semantic meaning.
\item We design a \textbf{seasonal-trend correction (STCorrection)} module to capture high-order dependencies and ensure coherent alignment between components, addressing the limitations of marginally independent modeling and enhancing temporal consistency of generated sequences.
\item We demonstrate that STDiffusion achieves \textbf{state-of-the-art performance} across diverse benchmarks, and generalizes well to \textbf{long-sequence generation tasks}, showcasing its robustness and scalability.
\end{itemize}

\section{Related Work}
\subsection{GAN-based Works}
GAN-based models are crucial in many early stage generation tasks. C-RNN-GAN \cite{b53} was one of the earliest works to apply GANs to sequential data, using recurrent neural networks for both the generator and the discriminator. RCGAN \cite{b4} employs recurrent neural networks (RNNs) in both the generator and the discriminator for sequential data generation, with the conditional variant incorporating auxiliary information to guide the process, making it ideal for context-sensitive tasks. Not only relies solely on adversarial feedback, TimeGAN \cite{b5} introduces a supervised loss to explicitly model stepwise temporal relationships, ensuring that generated sequences closely align with the conditional distributions in the training data. COT-GAN \cite{b6} optimizes the generator with an adversarial loss based on causal optimal transport, integrating optimal transport with temporal causality constraints, making it ideal for time-dependent data distributions.

\subsection{VAE-based Works}
VAE-based models are alternative generation models with comparable performance and greater efficiency. TimeVAE \cite{b9} provides a stable alternative by modeling sequences with a Gaussian prior, with the ability to encode domain knowledge and incorporating trend and seasonality information into the decoder. CR-VAE \cite{b10} combines Granger causal graph learning with a multihead decoder and an error compensation module to account for residual influences, ensuring accurate modeling while uncovering causal relationships among the variables. KoVAE \cite{b11}, inspired by Koopman theory, uses a novel model prior with a linear map in the latent space to generate time series from both regular and irregular data.

\subsection{Diffusion-based Works}
Given the capability of diffusion models to address the challenging problem of high-quality time series generation, several advanced methods have been proposed recently. GuidedDiffTime \cite{b54} supports both soft and hard constraints during generation and eliminates the need for model retraining, enabling efficient and realistic synthesis of counterfactual time series in several domains. FTS-Diffusion \cite{b20} uses a diffusion-based network to synthesize financial time series by recognizing and generating irregular, scale-invariant patterns. Unlike traditional diffusion-based models, Diffusion-TS \cite{b21} employs an encoder-decoder transformer with disentangled temporal representations and a Fourier-based loss term to directly reconstruct the sample instead of the noise in each diffusion step.

\section{Methodology}
\begin{figure}[htbp]
\centerline{\includegraphics[scale=.7]{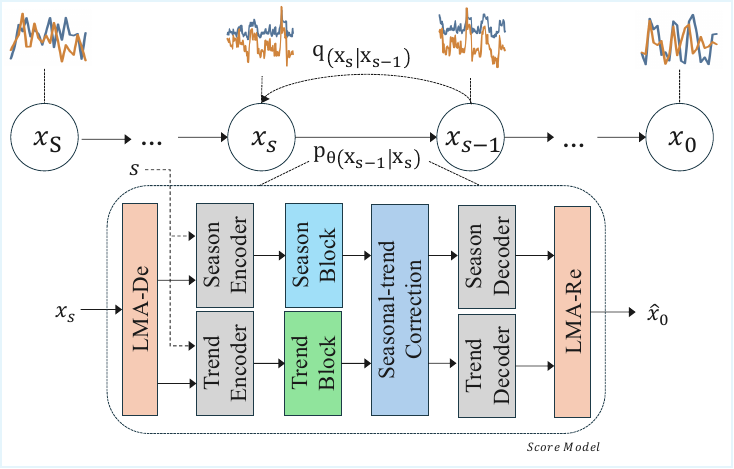}}
\caption{Illustrations of model structure consists of two parts: (a)A diffusion model schema involving add-noise and de-noise stages. (b) STDiffusion model essential components and pipeline.}
\label{fig: model strcture}
\end{figure}

In this section, we introduce the key components of STDiffusion shown in \autoref{fig: model strcture}, detailing their functional mechanisms and theoretical foundations. The pseudocode of the framework is described in Algorithm\autoref{algo: diffusion}. 

Our model is based on the denoising diffusion probabilistic framework \cite{b18} to generate realistic time series by explicitly modeling temporal components. The diffusion process consists of two phases: training and sampling. During the training phase, Gaussian noise is gradually added to a clean sequence $\mathbf{x}_0$, transforming it into a latent variable $\mathbf{x}_S \sim \mathcal{N}(\mathbf{0}, \mathbf{I})$ through a forward process $\mathbf{q}(\mathbf{x}_s \mid \mathbf{x}_{s-1})$. A score-based model is then trained to approximate the reverse denoising process $\mathbf{p}_\theta(\mathbf{x}_{s-1} \mid \mathbf{x}_s)$, using a noise estimator $\boldsymbol{\epsilon}_\theta$ to predict the injected noise at each step. The objective minimizes the discrepancy between the true and predicted noise:

\[\mathcal{L}(\mathbf{x}_0) := \min_\theta \mathbb{E}_{\mathbf{x}_0, s} \left\| \boldsymbol{\epsilon} - \boldsymbol{\epsilon}_\theta(\mathbf{x}_s, s) \right\|^2. \tag{1}\]

In the sampling phase, a latent $\mathbf{x}_S$ is drawn from a standard Gaussian distribution $\mathcal{N}(\mathbf{0}, \mathbf{I})$, and the learned reverse process $\mathbf{p_\theta}$ is applied iteratively to generate a sample $\mathbf{x}_0$ that resembles the original data distribution.

At each denoising step s, the model enhances temporal awareness by decomposing $\mathbf{x}_s$ using a Learnable Moving Average module (LMA-De), which yields trend and seasonal components, where season denotes the seasonal pattern plus residual error. These components are separately encoded and refined through dedicated Season and Trend Blocks and then reconciled via a Seasonal-Trend Correction module to capture their interaction. The adjusted representations are decoded and reassembled through LMA-Re to produce a denoised estimate $\hat{\mathbf{x}}_0$. During the sampling phase, the generation begins from pure Gaussian noise $\mathbf{x}_S \sim \mathcal{N}(\mathbf{0}, \mathbf{I})$, and the trained reverse process $\mathbf{p}_\theta(\mathbf{x}_{s-1} \mid \mathbf{x}_s)$, guided by the same component-aware architecture, is applied iteratively to reconstruct a realistic time series sample that reflects the underlying structural dynamics of the data.

\begin{algorithm}[tb]
\caption{Diffusion Framework}
\label{algo: diffusion}
\begin{algorithmic}[Diffusion Framework Pipeline]
\REQUIRE Data distribution $q(\mathbf{x}_0)$, steps $S$, schedule $\{\beta_s\}_{s=1}^S$
\textbf{Training Phase:}
\REPEAT
\STATE $\mathbf{x}_0 \sim q(\mathbf{x}_0)$, $s \sim \text{Uniform}({1,\ldots,S})$
\STATE $\boldsymbol{\epsilon} \sim \mathcal{N}(\mathbf{0},\mathbf{I})$
\STATE $\mathbf{x}_s = \sqrt{\bar{\alpha}_s}\mathbf{x}_0 + \sqrt{1-\bar{\alpha}_t}\boldsymbol{\epsilon}$
\STATE Update $\theta$ with $\nabla_{\theta} |\boldsymbol{\epsilon} - \boldsymbol{\epsilon}_{\theta}(\mathbf{x}_s, s)|^2$
\UNTIL{converged}

\REQUIRE Model parameters $\theta$, steps $S$, schedule $\{\beta_s\}_{s=1}^S$
\STATE \textbf{Sampling Phase:}
\STATE $\mathbf{x}_s \sim \mathcal{N}(\mathbf{0},\mathbf{I})$
\FOR{$s=S,\ldots,1$}
\IF{$s > 1$}
\STATE $\mathbf{z} \sim \mathcal{N}(\mathbf{0},\mathbf{I})$
\ELSE
\STATE $\mathbf{z} = \mathbf{0}$
\ENDIF
\STATE $\mathbf{x}_{s-1} = \frac{1}{\sqrt{\alpha_s}} \left(\mathbf{x}_s - \frac{\beta_s}{\sqrt{1-\bar{\alpha}_s}} \boldsymbol{\epsilon}_\theta(\mathbf{x}_s, s)\right) + \sigma_s\mathbf{z}$
\ENDFOR
\STATE \textbf{return} $\mathbf{x}_0$
\end{algorithmic}
\end{algorithm}

\subsection{Learnable Moving Average}
\begin{figure}[htbp]
\centerline{\includegraphics[scale=.7]{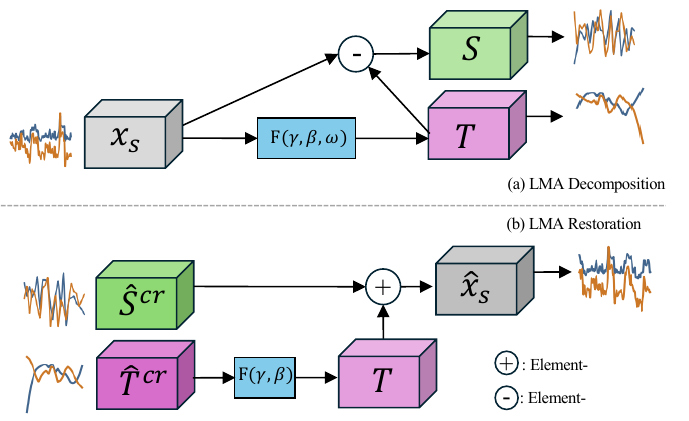}}
\caption{Learnable Moving Average pipeline: (a) upper part depicts the decomposition, and (b) lower part describes the restoration. $\oplus$ refers element-wise addition, and $\ominus$ refers element-wise deduction.}
\label{fig: LMA}
\end{figure}

One of our key contributions is the Learnable Moving Average (LMA), which adaptively balances the extraction of global trends with the preservation of local variations. Unlike traditional moving average methods \cite{b24, b31}, which rely on fixed-size average pooling windows and often blur important local structures across different datasets, LMA introduces a reversible, learnable mechanism that dynamically adjusts to the underlying data. This enables the model to retain meaningful local details while still capturing broader temporal patterns. The design of this mechanism is illustrated as $\mathbf{F(\gamma, \beta, \omega)}$ and $\mathbf{F(\gamma, \beta)}$ in \autoref{fig: LMA}. 

The LMA operates in two stages: decomposition and restoration, as illustrated in \autoref{fig: LMA}. During decomposition, the noisy input sequence $\mathbf{x}_s$ is separated into a trend component $\mathcal{T}$ and a seasonal component $\mathcal{S}$. In the restoration stage, these components are recombined to produce the final prediction.

\subsubsection{Decomposition}
The LMA begins with a latent input time series $\mathbf{x_S} \in \mathbb{R}^{L \times K}$, obtained by adding Gaussian noise to the original sequence $\mathbf{x_0} \in \mathbb{R}^{L \times K}$. To extract trend information, we compute multiple kernel-based moving averages using window sizes $l_k \in \{1, 2, 4, 6, 12\}$.  For the k-th moving average with window size $l_k$ in timestep $t$: $MA_k(t) = \frac{1}{l_k} \sum_{i=0}^{l_k-1} \mathbf{x_s}^{t-i}$. We combined all moving averages into a vector: $[\text{MA}_1(t), \text{MA}_2(t), ..., \text{MA}_5(t)]$, and an MLP followed by a softmax layer learns the weights $[w_1, w_2, ..., w_5]$, where $\sum_{i=1}^5 w_i = 1$. Subsequently, to capture local contextual information on trends, we apply an affine transformation with learnable scale $\gamma$ and bias $\beta$, which are stored and later reused in the restoration stage. This trend extraction process can be expressed as
\[\mathcal{T} = \mathbf{F(\gamma, \beta, \omega)}=\gamma \cdot \sum_{i=1}^N w_i \cdot (\frac{\sum_{j=0}^{l_i-1} \mathbf{x_s}^{t-j}}{l_i}) + \beta, \tag{2}\]
which corresponds to $\mathbf{F(\gamma, \beta, \omega)}$ in the upper part of \autoref{fig: LMA}. 

After obtaining the final trend component $\mathcal{T}$, we define the seasonality component $\mathcal{S}$ as the subtractive offset of the trend from the input $\mathbf{x_s}$.

\subsubsection{Restoration}
Restoration begins after the Seasonal-Trend Correction module (see Section D). The corrected trend $\mathcal{T}^{cr}$ and seasonality are first decoded to their original dimensions, after which LMA-Re applies the stored affine parameters to the trend and combines them with the seasonality to produce the final model output as follows:
\[\hat{\mathbf{x}}_0 =\mathbf{F(\gamma, \beta)}= \frac{\hat{\mathcal{T}}^{cr} - \beta}{\gamma} + \hat{\mathcal{S}}^{cr}, \tag{3}\] 
where $\mathbf{F(\gamma, \beta)}$ is also mentioned in the lower part of \autoref{fig: LMA}.

\subsection{Trend Learning Block (TLBlock)}
To capture trend patterns, we design a TLBlock that processes the extracted trend components. This block begins by applying instance normalization to mitigate distributional shifts and stabilize the input, then passes the normalized features through multiple residual-connected MLP layers, and finally denormalizes the outputs to restore the original scale. This architecture aligns with the findings of TDformer \cite{b28}, which show that MLPs are particularly effective in modeling trends, as they avoid the softmax-induced polarization effect in attention mechanisms that tends to overemphasize low-frequency signals. However, directly applying MLPs to raw trend signals can still be problematic under non-stationary conditions, where distribution shifts undermine model stability. To address this, we integrate reversible instance normalization (RevIN) \cite{b27} into our TLBlock to normalize such shifts during training while retaining the ability to reconstruct the true signal scale, ensuring robust and interpretable trend representation. The calculation in TLBlock is formulated as follows:
\[\hat{\mathcal{T}} = \mathbf{RevIN}_{re}\left(\sum_{i=1}^N \mathbf{MLP}\left(\mathbf{RevIN}_{norm}(\mathcal{T})\right)\right), \tag{4}\]
where $\mathbf{RevIN}_{re}$ and $\mathbf{RevIN}_{norm}$ denote restoration and normalization operations, respectively.

\subsection{Seasonal Learning Block (SLBlock)}
To model the seasonal component, we introduce a learnable wavelet transformation that aligns naturally with the multiscale and hierarchical nature of time series. This method has also demonstrated strong performance in tasks such as cosmology and molecular-partner prediction by effectively extracting distribution-specific features \cite{b26}. Our learnable wavelet enhanced with a learnable low-pass filter $\mathbf{h}_\theta$ is able to decompose seasonality according to the intrinsic pattern of the data. Unlike regular wavelet transformation that brutally decomposes along the middle frequency, our learnable wavelet decomposes seasonality according to the data intrinsic patterns, adaptively providing interpretable frequencies.  Intuitively, this allows the model to adaptively focus on frequency components that are most informative for the given data. The SLblock consists of three stages: learnable wavelet decomposition, frequency learning, and wavelet reconstruction. We use order-3 Daubechies wavelets (db3) \cite{b46} with periodic boundary conditions in our wavelet transformation setup.

\subsubsection{Learnable Wavelet Decomposition}
The seasonal input $\mathcal{S}$ is processed through a learnable low-pass filter $\mathbf{h}_\theta$. At each scale level $j$, the input approximation coefficients $a_j[n]$ are convolved with the low-pass filter $\mathbf{h}_\theta$ and the high-pass filter $\mathbf{g}$, followed by downsampling by a factor of 2. This yields the next-level approximation coefficients $a_{j+1}[p]$ and detail coefficients $d_{j+1}[p]$. 
\[\begin{cases}
\label{eq:wavelet}
a_{j+1}[p] = \sum_n \mathbf{h}_\theta[n-2p]a_j[n] = a_j \star {\mathbf{h}}_\theta[-2p] \\
d_{j+1}[p] = \sum_n \mathbf{g}[n-2p]a_j[n] = a_j \star {\mathbf{g}}[-2p]
\end{cases},\tag{5}\]
Specifically, $a_j \star \mathbf{h}_\theta[-2p]$ and $a_j \star \mathbf{g}[-2p]$ represent the strided convolutions that enable multi-resolution analysis. Unlike fixed filters in traditional wavelet transforms, the learnable $\mathbf{h}_\theta$ adapts to the data, allowing more flexible and data-driven decomposition.

In summary, the seasonal input $a_0 = \mathcal{S}$ is decomposed through a multilevel wavelet transform into a set of frequency components. At each level of decomposition $j$, the detail coefficient (high frequency) is denoted by $d_j$, and the approximation coefficient (low frequency) is denoted by $a_j$. After J levels of decomposition, we obtain a sequence of detail components $d_1, d_2, \ldots, d_J,$ along with a final approximation component $a_J$. For simplicity, we define a unified notation $\mathcal{F}_{L_i}^i$ to represent these components, where:
\begin{itemize}
    \item $\mathcal{F}_{L_i}^i = d_i$ for $i = 1, 2, \ldots, J $(detail frequencies),
    \item $\mathcal{F}_{L_{J+1}}^{J+1} = a_J$ (approximation frequency).
\end{itemize}

\subsubsection{Frequency Learning}
After we decompose seasonality into frequencies, we apply a self-attention mechanism \cite{b34} to model the interactions among these frequency patterns, with formulation as follows:
\[\hat{\mathcal{F}}_{L_{i}}^{\,i}
= \text{softmax}\left(\frac{Q^{^{\mathcal{F}^{i}}}(K^{^{\mathcal{F}^{i}}})^T}{\sqrt{d_k}}\right)V^{^{\mathcal{F}^{i}}}, 
\qquad i = 1, 2, \ldots
\tag{6}\]
where $\hat{\mathcal{F}}_{L1}^{i}$ denotes the predicted frequency of the $i^{th}$ level with length $Li$. $Q^{^{\mathcal{F}^{i}}}$, $K^{^{\mathcal{F}^{i}}}$, and $V^{^{\mathcal{F}^{i}}}$ are the three different linear transformations of $\mathcal{F}_{L_i}^i$, and $d_k$ is the dimension of $K^{^{\mathcal{F}^{i}}}$.

\subsubsection{Wavelet Reconstruction}
After obtaining the learned frequencies $\hat{\mathcal{F}}_{Li}^{i}$, we employ learnable wavelets to reconstruct these frequencies into the predicted seasonality $\hat{\mathcal{S}}$. Equation \eqref{eq:inverse_wavelet} defines the wavelet reconstruction process, where the signal on scale $j$, denoted by $\hat{a}_j[p]$, is recovered by combining the predicted approximation coefficients $\hat{a}_{j+1}[n]$ and the detail coefficients $\hat{d}_{j+1}[n]$ from the finer scale $j+1$. This is achieved through convolution with the learnable low-pass filter $\mathbf{h}_\theta$ and high-pass filter $\mathbf{g}$, followed by upsampling:
\begin{equation}
\label{eq:inverse_wavelet}
\hat{a}_j[p] = \sum_n \mathbf{h}_{\theta}[p-2n]\hat{a}_{j+1}[n] + \sum_n \mathbf{g}[p-2n]\hat{d}_{j+1}[n], \tag{7}
\end{equation}
where the $\mathbf{h}_\theta$ parameters are shared with those used in the decomposition stage.
Similarly, we define $\hat{\mathcal{F}}_{L_i}^i = \hat{d}_i$ for $i = 1, 2, \ldots$ and $\hat{\mathcal{F}}_{L_{J+1}}^{J+1} = \hat{a}_J$. The predicted seasonality $\hat{\mathcal{S}}$ corresponds to the reconstructed signal $\hat{a}_0$, which is obtained by inverting the wavelet transform using $\hat{a}_J$ and $\hat{d}_1, \ldots, \hat{d}_J$.

\subsection{Seasonal-Trend Correction}
After TLBlock and SLBlock produce their learned trends $\hat{\mathcal{T}}$ and seasonality $\hat{\mathcal{S}}$, we still need to further correct for their latent representations. 

We first project the predicted trend $\hat{\mathcal{T}}$ and seasonality $\hat{\mathcal{S}}$ into a double channel space, then divide it into two halves by operation $\mathbf{chunk}$: $\mathbf{chunk}: \mathbb{R}^{L \times 2d} \rightarrow \mathbb{R}^{L \times d} \times \mathbb{R}^{L \times d}$. Half serves as input, while the other functions as conditional term:
\[[\hat{\mathcal{T}}^{in}, \hat{\mathcal{T}}^{cnd}] = \text{chunk}(\mathbf{MLP}(\hat{\mathcal{T}})), \quad \hat{\mathcal{T}}^{in}, \hat{\mathcal{T}}^{cnd} \in \mathbb{R}^{L \times d}\]
\[[\hat{\mathcal{S}}^{in}, \hat{\mathcal{S}}^{cnd}] = \text{chunk}(\mathbf{MLP}(\hat{\mathcal{S}})), \quad \hat{\mathcal{S}}^{in}, \hat{\mathcal{S}}^{cnd} \in \mathbb{R}^{L \times d} \tag{8}\]
where $\hat{\mathcal{T}}^{in}$ and $\hat{\mathcal{S}}^{in}$ represent input terms; $\hat{\mathcal{T}}^{cnd}$ and $\hat{\mathcal{S}}^{cnd}$ denote conditional terms.

\begin{figure}[htbp]
\centerline{\includegraphics[scale=0.95]{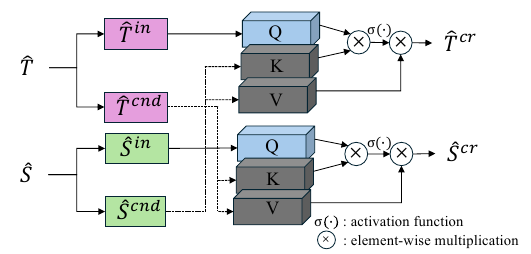}}
\caption{Seasonal-Trend Correction module. Predicted trend and seasonality are refined via cross-attention between input and conditional branches.}
\label{fig: STCorrection}
\end{figure}

To perform a correction, we adopt the attention-based interaction mechanism proposed in \cite{b34}, as illustrated in \autoref{fig: STCorrection}. Specifically, the predicted trend and seasonality from the input stream, $\hat{\mathcal{T}}^{in}$ and $\hat{\mathcal{S}}^{in}$, are projected into the query space Q, while the corresponding components of the conditional stream, $\hat{\mathcal{T}}^{cnd}$ and $\hat{\mathcal{S}}^{cnd}$, are used to compute the keys K and the values V, which are performed as:
\[\hat{\mathcal{T}}^{cr} = \mathbf{A}(Q^{\hat{\mathcal{T}}^{in}}, K^{\hat{\mathcal{S}}^{cnd}}, V^{\hat{\mathcal{S}}^{cnd}}), \tag{9}\]
\[\hat{\mathcal{S}}^{cr} = \mathbf{A}(Q^{\hat{\mathcal{S}}^{in}}, K^{\hat{\mathcal{T}}^{cnd}}, V^{\hat{\mathcal{T}}^{cnd}}), \tag{10}\]
with linear transformations: $Q^{\hat{\mathcal{T}}^{in}} = \hat{\mathcal{T}}^{in}W_Q$, $K^{\hat{\mathcal{S}}^{cnd}} = \hat{\mathcal{S}}^{cnd}W_K$, $V^{\hat{\mathcal{S}}^{cnd}} = \hat{\mathcal{S}}^{cnd}W_V$, and similarly for the seasonal component. Here, $W_Q$, $W_K$, and $W_V$ are learnable projection matrices, and $d_k$ denotes the dimensionality of the key vectors. The corrected representations $\hat{\mathcal{T}}^{cr}$ and $\hat{\mathcal{S}}^{cr}$ are then forwarded to the LMA-Re module for final reconstruction.

\section{Experiment}

\subsection{Experiment Setting}

\subsubsection{Baselines}
We compare STDiffusion with six baseline models that span three categories: GAN-based models including TimeGAN \cite{b5} and Cot-GAN \cite{b6}, VAE-based approaches including TimeVAE \cite{b9} and KoVAE \cite{b11}, and diffusion-based methods including Diffusion-TS \cite{b21} and Diffwave \cite{b45}. These selections provide a thorough evaluation across different methodological approaches to time series generation. We also select eight real-world datasets elaborated in \autoref{tab:dataset_info} for an in-depth comparison among the models.

\begin{table}[htbp]
\centering
\caption{Summary of datasets used in experiments}
\begin{tabular}{lccc}
\hline
\textbf{Dataset} & \textbf{Timestamps} & \textbf{Features} & \textbf{Interval} \\
\hline
ETTh1\cite{b40} & 17,520 & 8 & 1 hour \\
ETTh2\cite{b40} & 17,520 & 8 & 1 hour \\
Energy\cite{b41} & 19,735 & 28 & 10 minutes \\
fMRI\cite{b52} & 10000& 50&  2 seconds \\
Stock\cite{b51} &3773\textsuperscript{a} &7 & 1 day \\
Exchange\cite{b42} & 9,496\textsuperscript{a} & 8 & 1 day \\
Weather\cite{b43} & 52,560\textsuperscript{a} & 21 & 10 minutes \\
Occupancy\cite{b44} & 10,129 & 18 & 30 seconds \\
\hline 
\multicolumn{4}{l}{\textsuperscript{a}Calculated based on given date ranges} \\
\end{tabular}
\label{tab:dataset_info}
\end{table}

\begin{table*}[htbp]
\caption{Standard Time Series Generation Score with a generated sequence length of 24. Each score is the average with a 95\% confidence interval of 5 evaluation trials. The best results are in bold, and the second best are underlined. All scores are lower, the better}
\begin{center}
\renewcommand{\arraystretch}{1.1}
\begin{tabular}{c|c|*{8}{c}}

\hline
\multirow{2}{*}[0ex]{\parbox{2cm}{\centering \textbf{Metric}}} & \multicolumn{9}{c}{\textbf{Performance Measures}} \\
\cline{2-10}
& \multicolumn{1}{c|} {\textbf{Model}} & \textbf{ETTh1} & \textbf{ETTh2} & \textbf{Energy} & \textbf{fMRI} & \textbf{Stock} & \textbf{Exchange} & \textbf{Weather} & \textbf{Occupancy}\\
\hline
\multirow{7}{*}[0ex]{\parbox{2cm}{\centering Discriminative Score}$\big\downarrow$}
& \centering STDiffusion & \textbf{0.025±.010} &  \textbf{0.016±.010} & \textbf{0.107±.011} & \textbf{0.157±.046} & \textbf{0.006±.005} & \textbf{0.004±.003} & \textbf{0.071±.005} & \textbf{0.066±.026}\\
& \centering Diffusion-TS & \underline{0.061±.009} & \underline{0.038±.005} & \underline{0.122±.009} & \underline{0.167±.023} & \underline{0.067±.015} & \underline{0.034±.019} & \underline{0.161±.007} & \underline{0.120±.025} \\
& \centering TimeGAN  & 0.114±.055 & 0.094±.014 & 0.236±.012 & 0.484±.042 & 0.102±.021 & 0.275±.060  & 0.378±.042  & 0.365±.014  \\
& \centering TimeVAE & 0.244±.055 & 0.099±.075 & 0.499±.000 & 0.476±.044 & 0.145±.120 & 0.106±.043  & 0.391±.022  & 0.499±.023 \\
& \centering KoVAE & 0.201±.022 & 0.140±.064 & 0.373±.050 & 0.470±.046 & 0.124±.085 & 0.082±.043  & 0.394±.017  & 0.324±.035 \\
& \centering Diffwave & 0.228±.008 & 0.374±.009 & 0.479±.005 & 0.402±.029 & 0.232±.061 & 0.318±.009  & 0.497±.000  & 0.333±.025\\
& \centering Cot-GAN & 0.325±.099 & 0.499±.000 & 0.498±.002 & 0.492±.002 & 0.230±.016 & 0.498±.001  & 0.491±.002  & 0.436±.138 \\
\hline
\multirow{8}{*}[0ex]{\parbox{2cm}{\centering Predictive Score}$\big\downarrow$}
& \centering STDiffusion & \textbf{0.117±.006} &  \textbf{0.105±.002} & \textbf{0.250±.000} & \textbf{0.099±.000} & \textbf{0.036±.000} & \textbf{0.034±.002}  & \textbf{0.001±.000}  & \textbf{0.009±.000} \\
& \centering Diffusion-TS & \underline{0.119±.002} & \underline{0.107±.004} & \underline{0.250±.009} & \underline{0.099±.000} & \underline{0.036±.000} & \underline{0.034±.002}  & \underline{0.001±.000} & \underline{0.009±.000} \\
& \centering TimeGAN & 0.124±.001 &	0.118±.004 & 0.273±.004 & 0.126±.002 & 0.038±.001 & 0.066±.002 & 0.002±.000 & 0.057±.001\\
& \centering TimeVAE & 0.134±.002 & 0.116±.001 & 0.277±.000 & 0.113±.003 & 0.039±.000 &	0.038±.001 & 0.002±.000 & 0.023±.002\\
& \centering KoVAE & 0.126±.003&	0.118±.006&	0.251±.000	& 0.175±.007 &	0.056±.008&	0.040±.000	&0.002±.000&	0.014±.001\\
& \centering Diffwave& 0.127±.004&	0.147±.006&	0.252±.000	&0.101±.000	&0.047±.000	&0.068±.011&	0.005±.000	&0.042±.002\\
& \centering Cot-GAN &0.129±.000&	0.354±.050&	0.259±.000&	0.185±.003&	0.047±0.00&	0.042±.211&	0.003±.000&	0.038±.002\\
\cline{2-10}
& \centering Original & 0.121±.005&	0.108±.005&	0.250±.003&	0.090±.001&	0.036±.001&	0.035±.003&	0.001±.000&	0.012±.005\\
\hline
\multirow{7}{*}[0ex]{\parbox{2cm}{\centering Context-FID Score}$\big\downarrow$}
& \centering STDiffusion & \textbf{0.068±.004} & \textbf{0.038±.006} & \textbf{0.081±.015} & \underline{0.231±.023} & \textbf{0.014±.002} & \textbf{0.014±.001} & \textbf{0.072±.010} & \textbf{0.061±.006}\\
& \centering Diffusion-TS & \underline{0.116±.010} & \underline{0.053±.009} &  \underline{0.089±.024} & \textbf{0.105±.006} & 0.147±.025 & \underline{0.053±.006} & 0.339±.046 & 0.660±.099\\
& \centering TimeGAN & 0.300±.013&	0.102±.017&	0.767±.103&	1.292±.218&	0.103±.013&	0.359±.068& \underline{0.310±.059} & 0.762±.240 \\
& \centering TimeVAE & 0.788±.115&	0.436±.025&	1.790±.091&	14.449±.969&	0.215±.035&	0.254±.050&	0.527±.065&	0.891±.146  \\
& \centering KoVAE & 1.508±.206&	0.263±.017&	0.263±.017& 1.761±.158	&	\underline{0.056±.008}&	0.175±.029&	0.432±.025&	\underline{0.494±.040}  \\
& \centering Diffwave & 1.649±.191&	7.081±.866&	4.175±.593&	0.244±.018&	0.232±.032&	4.094±.504&	3.047±.190&	0.991±.133  \\
& \centering Cot-GAN & 0.980±.071&	1.250±.020&	1.039±.028&	7.813±.550&	0.408±.086&	4.300±.045&	2.420±.040&	1.391±.230  \\
\hline
\multirow{7}{*}[0ex]{\parbox{2cm}{\centering Correlation Score}$\big\downarrow$}
& \centering STDiffusion & \textbf{0.035±.009} & \textbf{0.054±.009}  & \textbf{0.802±.170} & \textbf{0.912±.035} & \underline{0.009±.005} &\textbf{0.043±.019} &\textbf{0.675±.138} &\textbf{0.543±.046}\\
& \centering Diffusion-TS & \underline{0.049±.008} & \underline{0.069±.015} & \underline{0.856±.147} & \underline{1.411±.042} & \textbf{0.004±.001} &\underline{0.074±.026} & 1.388±.040 &\underline{0.868±.030} \\
& \centering TimeGAN & 0.210±.006&	0.135±.008&	4.010±.104&	23.503±.039&	0.063±.005&	0.217±.013&	1.916±.055&	1.032±.054  \\
& \centering TimeVAE & 0.104±.013&	0.093±.014&	2.136±.192&	17.296±.526&	0.095±.008&	0.135±.025&	0.970±.096&	2.343±.165 \\
& \centering KoVAE & 0.104±.012&	0.240±.017&	0.240±.017&  7.475±.044 & 0.060±.011 &	0.142±.025 & \underline{0.687±.109} & 1.599±.195\\
& \centering Diffwave & 0.191±.013&	0.579±.015&	5.360±.077&	3.927±.049&	0.030±.020&	0.611±.022&	1.445±.112&	1.253±.165 \\
& \centering Cot-GAN & 0.249±.009&		0.110±.009&	3.164±.061&	26.824±.449&	0.087±.004&	0.183±.037&	1.450±.060&	1.373±.085 \\
\hline
\end{tabular}
\label{tab:results}
\end{center}
\end{table*}

\subsection{Metric Definition}

\subsubsection{Visualization} We use three complementary methods to analyze and visualize time series patterns. \textbf{Principal Component Analysis (PCA)} \cite{b35} captures global temporal trends by projecting data onto directions of maximum variance. \textbf{t-Distributed Stochastic Neighbor Embedding (t-SNE)} \cite{b36} highlights local structure and clustering by minimizing divergence between high- and low-dimensional probability distributions. \textbf{Data Density Estimation} with Gaussian kernels \cite{b37} provides smoothed probability densities for direct comparison of distributions. Together, these methods offer insights into global variance, local clustering, and marginal distributions of real and generated data.

\subsubsection{Evaluation Metrics}
To comprehensively evaluate generation quality, we employ four metrics that assess different aspects of the synthetic time series: 
\begin{itemize}
\item \textbf{Discriminative Score} \cite{b5} evaluates distribution similarity using a GRU classifier to distinguish real from generated sequences. A score of 0 indicates perfect generation, while 0.5 suggests the inability to produce meaningful sequences. 
\item \textbf{Predictive Score} \cite{b5} evaluates temporal dynamics via one-step-ahead predictions, with a GRU predictor trained on synthetic data and tested on real data. A score of 0 indicates perfect generation, while 0.5 suggests the inability to produce meaningful sequences. 
\item \textbf{Context-FID} \cite{b38} measures the distribution distance in a learned embedding space, where the lower scores represent a higher similarity between real and synthetic data. 
\item \textbf{Correlational Score} \cite{b39} assesses temporal dependencies through cross-autocorrelation, where lower scores indicate better preservation of temporal dependencies.
\end{itemize}

\subsection{Standard Time Series Generation Results}
\autoref{tab:results} presents a comprehensive evaluation of our STDiffusion against eight baselines in four key metrics, and our model achieves superior performance in generating high-quality 24-length time series. For example, STDiffusion reduces the discriminative score by approximately 52\% relative to the next-best model and maintains the first place in the most predictive, Context-FID, and correlation scores. More specifically, in low-signal, high-noise datasets such as Stocks and Exchanges, where the improvement reaches more than 85\%. The method remains robust in high-dimensional datasets such as energy and fMRI, where it still secures the best (or second-best) figures in all four metrics, underscoring its ability to model complex multivariate dynamics. 

\begin{figure*}[htbp]
\centerline{\includegraphics[scale=1.4]{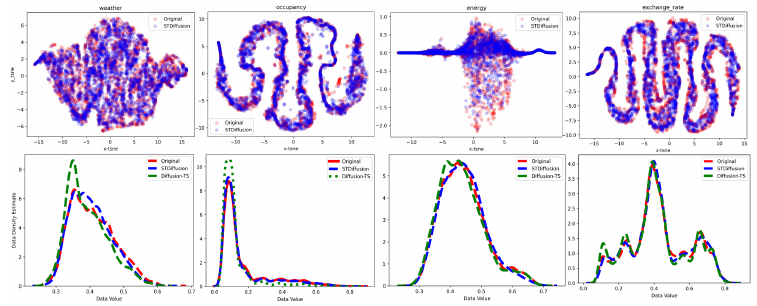}}
\caption{t-SNE visualizations (top row) compare the original data (red dots) with samples generated by STDiffusion (blue dots). The bottom row presents data density estimates, showing how the distributions of STDiffusion (blue lines) and Diffusion-TS (green lines) align with the ground truth (red lines).}
\label{fig: t-SNE}
\end{figure*}

In the more intuitive visualization analysis shown in \autoref{fig: t-SNE}, STDiffusion shows exceptional performance in capturing both local and global patterns. The t-SNE plots reveal remarkable preservation of local temporal structures, with generated sequences forming well-aligned clusters that closely match the real data patterns. The data density estimation plots further validate our model's state-of-the-art capability in emulating global data distributions, showing particularly impressive fits for challenging high-dimensional datasets like Weather (with complex meteorological patterns) and Occupancy (with multiple sensor measurements). This superior performance on large-scale, high-dimensional data underscores STDiffusion's scalability and robustness in handling complex real-world time series data.

\begin{figure*}[htbp]
\centerline{\includegraphics[scale=0.7]{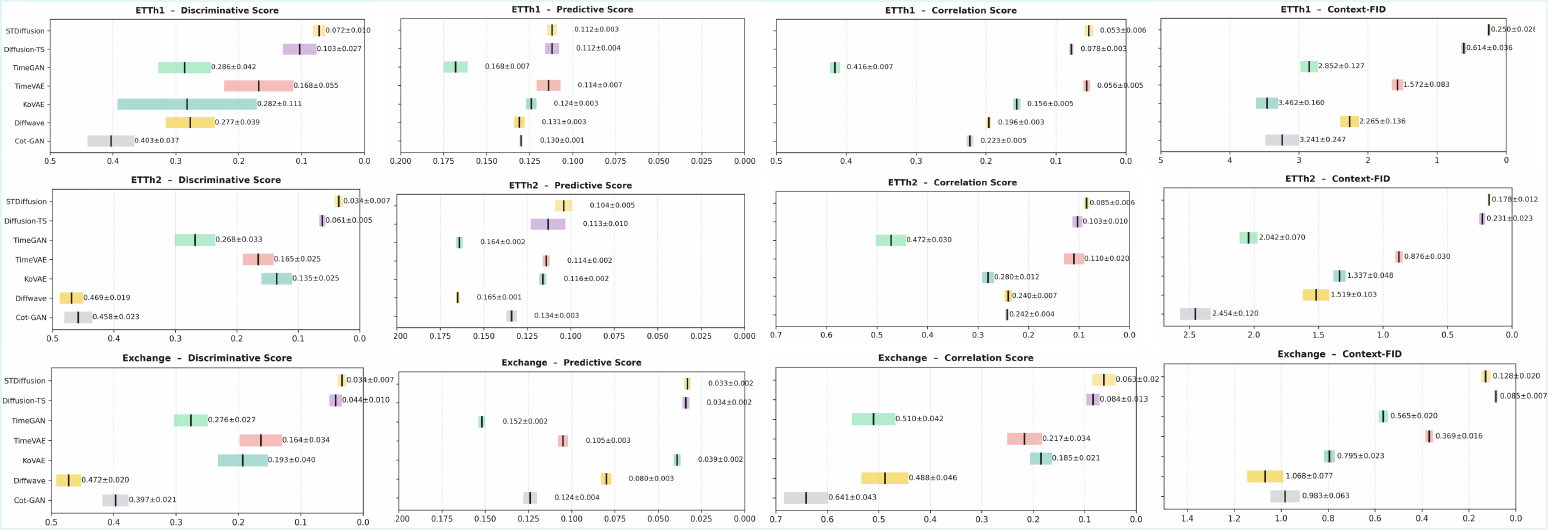}}
\caption{Results on ETTh1, ETTh2, and Exchange datasets. Bars show the average scores with 95\% confidence intervals for sequence lengths 64, 128, and 256. Lower values indicate better performance across all metrics. Additional score details is provided on the Github Repo \url{https://github.com/mobkageyama/STDiffusion}.}
\label{fig: longterm_generation}
\end{figure*}
\subsection{Long Sequence time series generation result}
Long sequence time series generation is crucial for various real-world applications, such as energy consumption forecasting, financial market simulation, and climate prediction, where the ability to generate coherent sequences over extended periods is essential for planning and risk assessment \cite{b47}\cite{b48}\cite{b49}. However, maintaining temporal consistency while preserving complex dependencies becomes increasingly challenging as the sequence length grows. To assess the robustness and long-term generation capability of various models under this constraint, we conducted a comprehensive evaluation across three challenging datasets: ETTh1, ETTh2 and Exchange. These datasets are characterized by complex temporal structures, including long-short range trends and strong seasonal effects, making them ideal for benchmarking long-sequence modeling.

We generate sequence lengths of 64, 128 and 256 with three datasets and evaluate them with the four previous metrics. \autoref{fig: longterm_generation} records each metrics average score of three different sequence lengths. STDiffusion keeps its lead in almost every evaluation of the long sequences generation. In addition to the mean score predominance, the confidence interval is also narrower than all other models, implying better stability and robustness of our model. Even on the most volatile Exchange dataset, it matches the second-best model in predictive score and sits within 0.04 of the context-FID. This indicates that STDiffusion is not only strong in modeling structured periodic data but also robust to noisy and less predictable sequences.

\subsection{Learnable Wavelet Function Visualization}
\begin{figure}[htbp]
\centerline{\includegraphics[scale=.28]{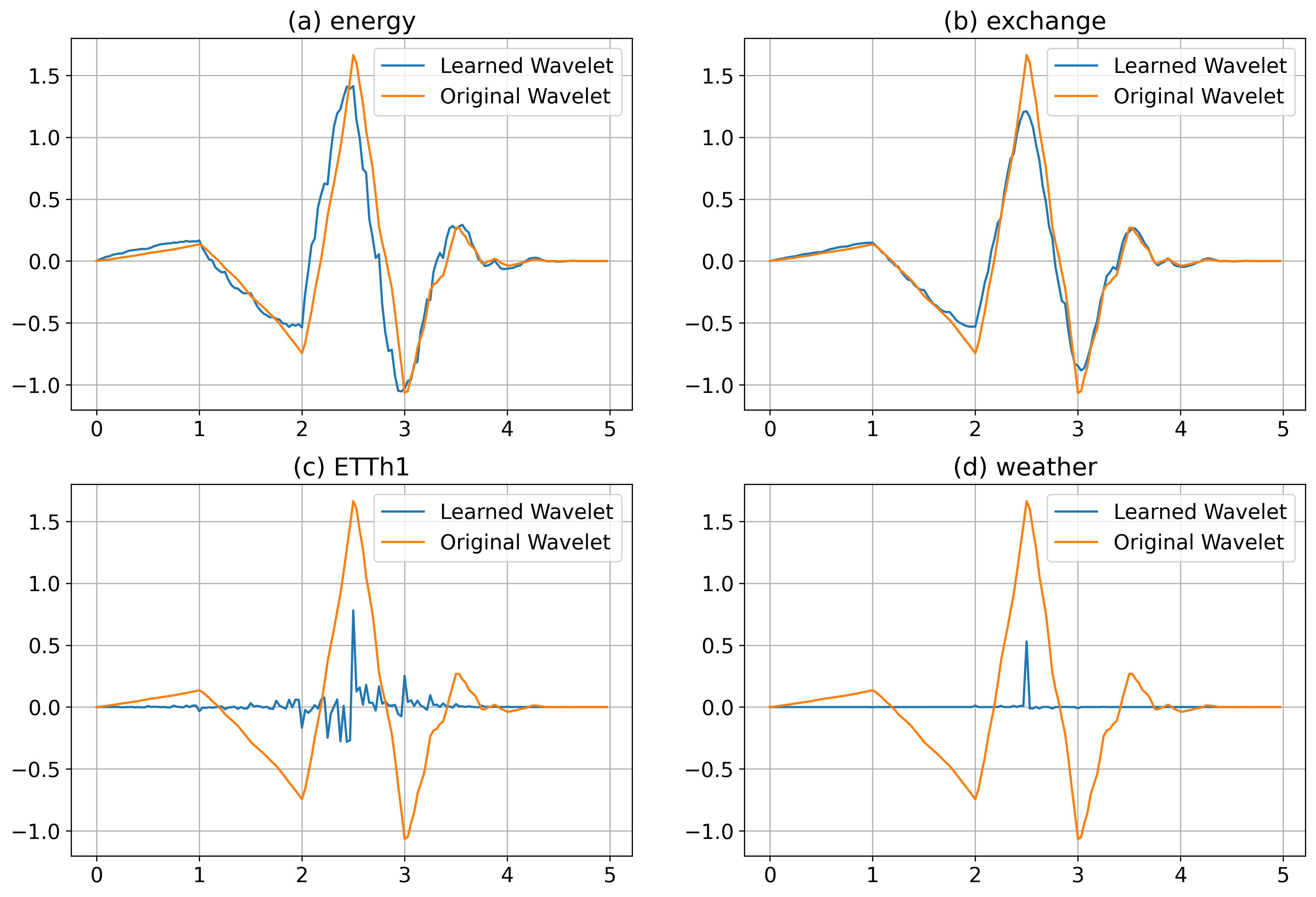}}
\caption{Visualization of learned wavelet functions for Energy, Exchange, ETTh1, and Weather datasets. The orange line shows the original db3 wavelet, while the blue line represents the learned wavelet adapted to each dataset.}
\label{fig: wavelet visualization}
\end{figure}

The visualization in \autoref{fig: wavelet visualization} shows approximations of wavelet functions determined by the low-pass filter coefficients $\mathbf{h}$ and wavelet type under the order-3 Daubechies wavelets (db3) \cite{b46} setting. The orange lines represent standard db3 wavelet with fixed coefficients, known for their asymmetry and three vanishing moments, which allow them to model polynomial signal structures. In contrast, the blue lines illustrate wavelet functions derived from our learned filter coefficients $\mathbf{h}_\theta$, which have been adapted to different data distributions. During training, $\mathbf{h}_\theta$ is constrained to match the length and orthogonality properties of standard db3 filters, ensuring that the resulting basis remains a valid wavelet frame tailored to each dataset’s frequency distribution.

\begin{table*}[htbp]
\centering
\caption{Ablation study for key components. Each score is the average with a 95\% confidence interval of 5 evaluation trials. For all metrics, lower values indicate better performance, and the best results are in bold}
\setlength{\tabcolsep}{5pt}
\begin{tabular}{l|l|ccccc}
\hline
\multicolumn{1}{c|}{\textbf{Metric}} &
\multicolumn{1}{c|}{\textbf{Method}} &
\textbf{ETTh1} & \textbf{ETTh2} & \textbf{Energy} & \textbf{fMRI} & \textbf{Exchange} \\
\hline
\multirow{4}{*}{\parbox{3cm}{Discriminative Score $\big\downarrow$}} 
  & STDiffusion              & \textbf{0.025±.010} & \textbf{0.016±.010} & \textbf{0.107±.011} & \textbf{0.157±.046} & \textbf{0.004±.003} \\
  & w/o Learnable Wavelet    & 0.026±.012 & 0.020±.010 & 0.107±.013 & 0.160±.055 & 0.004±.006 \\
  & w/o STCorrection         & 0.030±.012 & 0.022±.011 & 0.114±.014 & 0.158±.055 & 0.005±.006 \\
  & w/o LMA                  & 0.031±.013 & 0.022±.012 & 0.120±.016 & 0.200±.060 & 0.005±.007 \\
\hline
\multirow{4}{*}{\parbox{3cm}{Predictive Score $\big\downarrow$}} 
  & STDiffusion              & \textbf{0.117±.006} & \textbf{0.105±.002} & \textbf{0.250±.000} & \textbf{0.099±.000} & \textbf{0.034±.002} \\
  & w/o Learnable Wavelet    & 0.118±.007 & 0.112±.004 & \textbf{0.250±.000} & 0.105±.002 & \textbf{0.034±.002} \\
  & w/o STCorrection         & 0.120±.007 & 0.113±.004 & 0.252±.007 & 0.106±.003 & 0.043±.001 \\
  & w/o LMA                  & 0.121±.008 & 0.114±.005 & 0.252±.008 & 0.107±.004 & 0.044±.002 \\
\hline
\multirow{4}{*}{\parbox{3cm}{Context-FID $\big\downarrow$}} 
  & STDiffusion              & \textbf{0.068±.004} & \textbf{0.038±.006} & \textbf{0.081±.015} & \textbf{0.231±.023} & \textbf{0.014±.001} \\
  & w/o Learnable Wavelet    & \textbf{0.068±.004} & 0.043±.006 & 0.082±.011 & 0.265±.027 & \textbf{0.014±.001} \\
  & w/o STCorrection         & 0.073±.006 & 0.046±.007 & 0.095±.019 & 0.250±.026 & 0.017±.004 \\
  & w/o LMA                  & 0.075±.008 & 0.045±.007 & 0.089±.017 & 0.290±.032 & 0.019±.004 \\
\hline
\multirow{4}{*}{\parbox{3cm}{Correlation Score $\big\downarrow$}} 
  & STDiffusion              & \textbf{0.035±.009} & \textbf{0.054±.009} & \textbf{0.802±.170} & \textbf{0.912±.035} & \textbf{0.043±.019} \\
  & w/o Learnable Wavelet    & 0.040±.009 & 0.058±.009 & 0.820±.185 & 0.930±.040 & 0.044±.020 \\
  & w/o STCorrection         & 0.042±.009 & 0.065±.009 & 0.830±.180 & 0.940±.045 & 0.050±.016 \\
  & w/o LMA                  & 0.040±.010 & 0.070±.010 & 0.860±.190 & 0.960±.050 & 0.049±.026 \\
\hline
\end{tabular}
\label{tab:ablation_results}
\end{table*}

Across different datasets, these learned wavelets exhibit distinct adaptations that mirror the dominant spectral characteristics of each time series. In the Exchange dataset, where seasonality is predominantly low‐frequency and relatively stable, the learned wavelets assume a more pronounced low‐pass profile, yielding a smoother basis compared to the standard db3. This increased smoothness enables the model to capture gradual market cycles without exaggerating transient noise. In the ETTh1 dataset, which represents electricity transformer load with frequent periodic spikes and higher‐frequency fluctuations, the optimized wavelets show greater oscillatory behavior to reflect those rapid variations. The Weather dataset induces the most extreme adaptation: abrupt temperature or precipitation events create sharp discontinuities, and accordingly the learned wavelets suppress small‐scale oscillations almost entirely while retaining the ability to react to sudden spikes. In each case, the learned coefficients $\mathbf{h}_\theta$ reallocate the emphasis to the most informative frequency bands, thus improving both the interpretability and the fidelity of the seasonal decomposition.

\begin{figure}[htbp]
\centerline{\includegraphics[scale=.7]{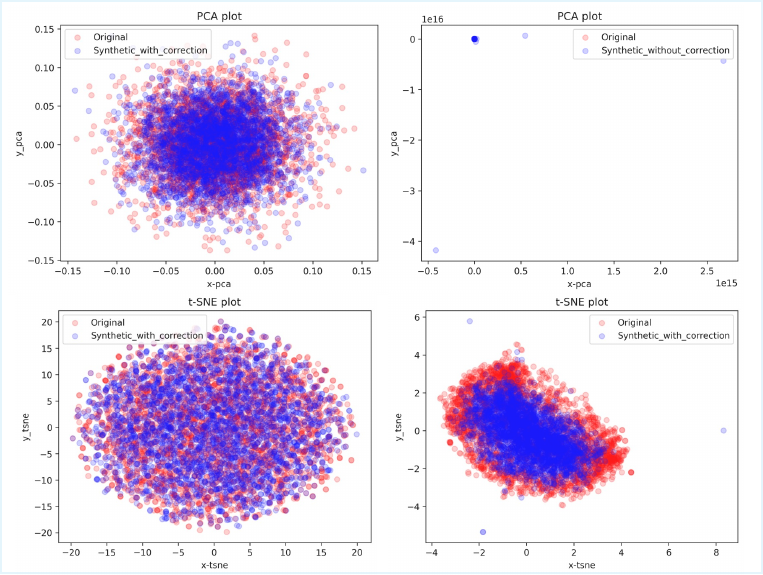}}
\caption{Comparison of ETTh1 trends and seasonality with and without correction. Red dots are original; blue dots are synthetic. The first row shows PCA results, and the second row shows t-SNE. Left: with correction. Right: without correction. }
\label{fig: ST_correct}
\end{figure}
\subsection{Ablation Study}
\autoref{tab:ablation_results} presents an ablation study across five datasets, evaluating discriminative, predictive, Context‐FID, and correlation metrics. Across nearly every metric and dataset, the full STDiffusion model achieves the best performance. In case of omitting LMA component, we replace it with a regular single kernel moving average with kernel size of 3. Omitting the LMA component produces the largest degradation, particularly noticeable in discriminative scores for ETTh1, ETTh2, Energy, and fMRI, as well as in both correlation and Context‐FID metrics on Exchange. We use a standard db3 wavelet transformation to replace the learnable wavelet and causes clear drops in discriminative and predictive metrics—this effect is most pronounced on ETTh2. Excluding the ST‐correction module by directly decoding predicted seasonality and trends from SLBlock and TLBlock also degrades generation quality, especially in Context‐FID and predictive scores across ETTh1, ETTh2, and fMRI. Interestingly, for energy and exchange predictive scores, two model configurations tie for the top performance. This outcome likely arises because, for both the Energy and Exchange datasets, the spectral characteristics align closely with those captured by the fixed db3 filter. These results confirm that each module—LMA, learnable wavelet, and ST‐correction—contributes meaningfully to STDiffusion’s success, and their combination yields the strongest, most consistent results across varied time series domains.

To further evaluate the impact of the Seasonal-Trend Correction module, we visualize the distributional alignment between real and synthetic samples using both PCA and t-SNE projections in the ETTh1 dataset, as shown in \autoref{fig: ST_correct}. The top row presents PCA plots, while the bottom row shows t-SNE plots. In both cases, red points represent real samples and blue points indicate generated sequences. On the left, the model with correction produces synthetic data that is well-aligned with the original distribution across both projections, showing significant overlap in both global and local structure. In contrast, the model without correction (right column) shows major divergence, particularly in the PCA space, where the display collapses due to the existence with extreme outliers in the synthetic samples. The t-SNE plot further reveals that the synthetic samples fail to capture the outlying pockets of the original distribution, instead collapsing toward the dense central region, along with several outliers. These results highlight the importance of the Seasonal-Trend Correction mechanism in stabilizing model performance.

\section{Discussion}

\subsection{LMA Integration in Forecasting Models}
We further evaluate the versatility of our LMA technique by applying it to two well-established forecasting models: Autoformer and FEDFormer. We replace the raw data decomposition and the final seasonal-trend combination stage used in the models with our LMA. We maintain the original hyperparameter settings for both models during the evaluation. The time series forecasting task involves predicting future values based on observed historical sequences, and its performance is commonly assessed using metrics such as Mean Squared Error (MSE) and Mean Absolute Error (MAE), where lower values indicate better predictive accuracy. As shown in \autoref{tab:forcast}, the integration of LMA into both models yields consistent improvements in the ETTh1 and Exchange datasets, particularly at longer horizons. For example, on Exchange with a length of 720, Autoformer+LMA reduces the MSE from 1.447 to 1.037 and the MAE from 0.941 to 0.788. On ETTh1, similar gains are observed across multiple settings. These results show that LMA not only strengthens generative capabilities, but also enhances temporal feature extraction in forecasting scenarios, confirming its effectiveness as a general-purpose, model-agnostic component for time series modeling.

\begin{table}[htbp]
\centering
\caption{Forecasting performance (MSE / MAE) on ETTh1 and Exchange. A lower MSE indicates better performance, and the best results are highlighted in bold.}
\small
\setlength{\tabcolsep}{2.7pt}

\begin{tabular}{c|c|cc cc cc cc}
\hline
      & \textbf{Models} &
        \multicolumn{2}{c}{\textbf{Autoformer}} &
        \multicolumn{2}{c}{\textbf{+LMA\textsuperscript{a}\rule{0pt}{2.2ex}}} &
        \multicolumn{2}{c}{\textbf{FEDFormer}} &
        \multicolumn{2}{c}{\textbf{+LMA\textsuperscript{b}}} \\
      & \textbf{Metrics} & MSE & MAE & MSE & MAE & MSE & MAE & MSE & MAE \\
\hline
\multirow{4}{*}{\rotatebox{90}{\textbf{ETTh1}}}
  &  96 & 0.449 & 0.459 & \textbf{0.436} & \textbf{0.446} & \textbf{0.374} & 0.414 & 0.375 & \textbf{0.413} \\
  & 192 & 0.500 & 0.482 & \textbf{0.452} & \textbf{0.458} & \textbf{0.427} & \textbf{0.448} & 0.431 & 0.451 \\
  & 336 & 0.521 & 0.496 & \textbf{0.513} & \textbf{0.494} & 0.457 & 0.467 & \textbf{0.455} & \textbf{0.464} \\
  & 720 & 0.514 & 0.512 & \textbf{0.488} & \textbf{0.489} & 0.506 & 0.507 & \textbf{0.495} & \textbf{0.502} \\
\hline
\multirow{4}{*}{\rotatebox{90}{\textbf{Exchange}}}
  &  96 & 0.197 & 0.323 & \textbf{0.155} & \textbf{0.290} & 0.148 & 0.278 & \textbf{0.144} & \textbf{0.277} \\
  & 192 & 0.300 & 0.369 & \textbf{0.266} & \textbf{0.377} & 0.271 & 0.380 & \textbf{0.265} & \textbf{0.378} \\
  & 336 & 0.509 & 0.524 & \textbf{0.439} & \textbf{0.492} & 0.460 & 0.500 & \textbf{0.430} & \textbf{0.485} \\
  & 720 & 1.447 & 0.941 & \textbf{1.037} & \textbf{0.788} & 1.195 & 0.941 & \textbf{1.133} & \textbf{0.807} \\
\hline
\multicolumn{10}{l}{\textsuperscript{a} Autoformer with LMA technique} \\
\multicolumn{10}{l}{\textsuperscript{b} FEDFormer with LMA technique} \\
\end{tabular}
\label{tab:forcast}
\end{table}

\subsection{Learnable Wavelet Parameters Study}
\begin{figure}[htbp]
\centerline{\includegraphics[scale=.23]{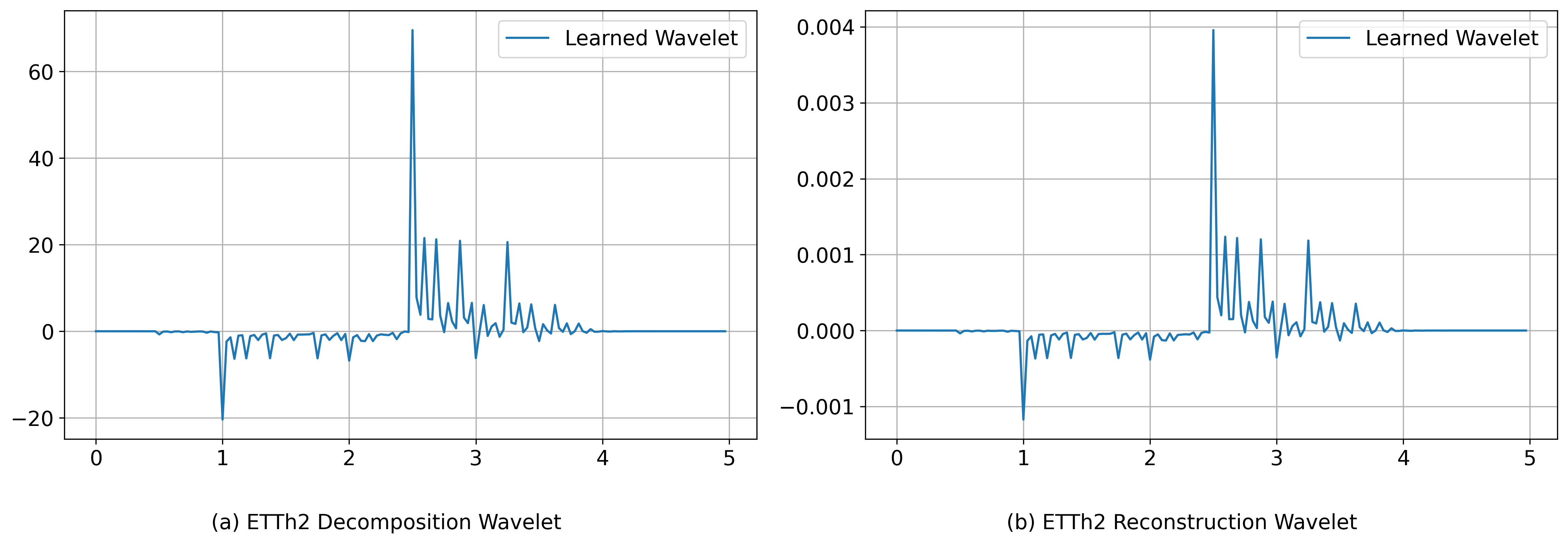}}
\caption{Wavelets Function in Decomposition and Reconstruction }
\label{fig: wavelet_ablation}
\end{figure}

To investigate whether the wavelet transform stages benefit from specialized representations, we experimented with using separate learnable low-pass filters $\mathbf{h}_\theta$ for the decomposition and reconstruction phases, rather than sharing a single set of parameters. Interestingly, we observed that these independent filters converge to remarkably similar patterns during training, as shown in \autoref{fig: wavelet_ablation}. However, while their shapes are nearly identical, there is a notable difference in their magnitude. The decomposition phase filters exhibited much higher magnitudes. According to \eqref{eq:wavelet}, the decomposition with high-magnitude filters will have strong high-frequency suppression since $\mathbf{h}_\theta$'s such filters will trim the high-frequency part. It causes overlooks in sharp transitions and edges, reducing the ability to detect sudden changes. In contrast, the reconstruction filters show notably smaller magnitudes. According to \eqref{eq:inverse_wavelet}, the reconstruction with low-magnitude filters will amplify the decomposed series since a lower filter amplitude allows broader frequency passage, introducing noise frequencies. Both phenomena can negatively impact the training stability; we therefore choose to use the shared $\mathbf{h}_\theta$ for both decomposition and reconstruction. 
\section{Conclusion}
In this paper, we have developed and demonstrated the efficacy of STDiffusion, a novel approach for time series generation that addresses the inherent challenges of capturing trends and seasonal patterns in complex datasets. Through a diffusion-based model, STDiffusion enhances the quality and realism of generated data by leveraging the gradual denoising process, which allows for the precise control of generation, resulting in a highly coherent data distribution. Our results in diverse datasets highlight the robustness and versatility of STDiffusion, particularly in scenarios with strong seasonality and high-dimensional data. The design of a learnable moving average and wavelet-based frequency analysis has further improved the model’s ability to generate accurate time series data. Additionally, the seasonal-trend correction layer plays a crucial role in aligning and maintaining the corresponding relationship between the generated trend and seasonal components, ensuring that the synthesized data remain true to the original distribution. Overall, this work contributes significantly to the field of time series generation, providing a powerful tool for both theoretical exploration and practical applications.

\end{document}